\documentclass[conference]{IEEEtran}
\IEEEoverridecommandlockouts
\usepackage{cite}
\usepackage{amsmath,amssymb,amsfonts}
\usepackage{algorithm, algorithmic}
\usepackage{graphicx}
\usepackage{textcomp}
\usepackage{xcolor}
\def\BibTeX{{\rm B\kern-.05em{\sc i\kern-.025em b}\kern-.08em
    T\kern-.1667em\lower.7ex\hbox{E}\kern-.125emX}}

\begin{document}
	\title{Deep Reinforcement Learning Based Dynamic Route Planning for Minimizing Travel Time}
	
	\author{\IEEEauthorblockN{Yuanzhe Geng, Erwu Liu, Rui Wang and Yiming Liu}
		
		\IEEEauthorblockA{\textit{College of Electronics and Information Engineering, Tongji University, Shanghai, China} \\
			Emails: yuanzhegeng@tongji.edu.cn, erwu.liu@ieee.org, ruiwang@tongji.edu.cn, ymliu\_970131@tongji.edu.cn}
	}
	
	\maketitle

\begin{abstract}
Route planning is important in transportation. 
Existing works focus on finding the shortest path solution or using metrics such as safety and energy consumption to determine the planning. 
It is noted that most of these studies rely on prior knowledge of road network, which may be not available in certain situations. 
In this paper, we design a route planning algorithm based on deep reinforcement learning (DRL) for pedestrians. 
We use travel time consumption as the metric, and plan the route by predicting pedestrian flow in the road network. 
We put an agent, which is an intelligent robot, on a virtual map. 
Different from previous studies, our approach assumes that the agent does not need any prior information about road network, but simply relies on the interaction with the environment. 
We propose a dynamically adjustable route planning (DARP) algorithm, where the agent learns strategies through a dueling deep Q network to avoid congested roads. 
Simulation results show that the DARP algorithm saves $52\%$ of the time under congestion condition when compared with traditional shortest path planning algorithms.
\end{abstract}

\begin{IEEEkeywords}
Route planning, pedestrian, deep reinforcement learning, minimum time consumption.
\end{IEEEkeywords}

\section{Introduction}\label{sect_intro}
As one of the most important parts of intelligent transportation system, route planning has been widely used in many areas, such as daily travels \cite{8405840, 7997454,7743527} and transportation and indoor warehouse management \cite{7988625,8126871,8577210}, so as to save time cost and management cost.

Most of existing studies are devoted to solving the shortest path problem without considering other metrics.
Traditional graph-based methods, such as Dijkstra and Floyd algorithms, have the serious issue of time complexity which will increase exponentially with the size of the graph.
Combining heuristic algorithm and graph-based methods, Hart \textit{et al.} proposed A* algorithm \cite{4082128} for better path searching by introducing an evaluation function, which is used to calculate the distance traveled to reach the current position and to predict the distance needed to travel in the future. Nowadays, many A* based algorithms, such as ARA*, Field D* and Block A* \cite{ARA,Dstar,BlockAstar}, have been widely applied in mobile robots and automatic vehicle navigation.
Unfortunately, because distance cannot reflect whether a road is congested, these methods cannot be applied directly to environment with traffic congestion.

In daily travel, people tent to choose less crowded road segments. Accordingly, we use total travel time as the optimization goal for route planning in this study.
In \cite{7815037}, Yan and Shen proposed a vehicle Trajectory based driving speed optimization strategy. Based on real-time vehicle density provided by central cloud server, each vehicle and the central server formed a non-cooperative Stackelberg game to minimize vehicle travel time.
In \cite{8789394}, Liu \textit{et al.} established a velocity-space-time three-dimensional network model by acquiring information such as real-time road traffic flow, traffic lights and vehicle running status, and realized the minimization of both time and energy consumption for the route.
Li \textit{et al.} \cite{LI2014936} studied the inventory path problem of large petroleum and petrochemical enterprise groups. Given distance and location information, they proposed an iterative heuristic algorithm to optimize total travel time while satisfying the constraints in enterprise scenarios.
In these studies, researchers consider the time cost, but they need to have prior knowledge of road network environment. In addition, these studies are for vehicles or automated machines, and are not directly applicable to navigation system for pedestrians.

To remove the dependency on prior knowledge of road environment, we employ reinforcement learning (RL) methods. RL is considered one of the three paradigms of machine learning, which is suitable for dealing with decision-making problems.
RL assumes that there is an intelligent entity called agent, which constantly interacts with the environment and adjusting its behavior strategy according to the feedback.
Traditional RL methods, such as Q-learning and Sarsa, use tables to store the returns that can be obtained by performing different actions in various states \cite{RL:intro2}. Therefore, when faced with high-dimensional state space, traditional methods often fail to work.
By combining deep neural network and RL algorithm, Mnih \textit{et al.} \cite{Mnih2015} proposed deep Q network (DQN), which can deal with highly dimensional space and has generalization ability. When the agent is in an unexperienced state, such DRL method can still make reasonable decisions.
Recently, some researches propose DRL based route planning systems, which take real-time weather, condition of wheels and road type into consideration, to optimize metrics such as safety and energy cost \cite{rps1,8456612,9028238}. These studies demonstrate that DRL is a promising technology in the field of intelligent transportation.

In this study we propose an agent-based approach for route planning. Unlike previous works, we assume that the agent does not have access to prior global information, and that all information is acquired by the agent's exploration.
The agent acts as a pedestrian, and explores its own way forward in road network based on local and partial knowledge, rather than being told what alternative paths are available. In order to enable the agent to effectively learn pedestrian traffic information in road network, we employ dueling deep Q network to train and improve the agent's behavior policy, so that the agent can select an optimal path with the minimum time consumption. Specifically, the contributions of this paper are summarized as follows.
\begin{itemize}
	\item
	We propose an autoregressive time series model to describe the relationship between pedestrian flow and time in each road segment. 
	With this model, one can accurately build a dynamic virtual environment for the training of the agent.
	
	\item
	We propose a DRL framework for route planning, where a reward function is designed to improve learning efficiency. While this study uses this framework for route planning that minimizes the travel time, one can also use it to solve route planning problems with other optimization objectives, such as the distance and safety.
	
	\item
	We propose a dynamically adjustable route planning (DARP) algorithm, which uses dueling deep Q network to learn from the environment. 
	By interacting with the environment, the agent can learn a behavior strategy based on real-time traffic data prediction, to minimize the total travel time from the origin to the destination.
	
\end{itemize}

The rest of this paper is organized as follows. Section \ref{sect rela} introduces the related works. Section \ref{sect model} analyzes the system model and Section \ref{sect problem} defines the route planning problem. Section \ref{sect DRL} presents our DRL based algorithm, and Section \ref{sect result} presents simulation results. Finally, we conclude this paper in Section \ref{sect Conclusion}.

\section{Related work}\label{sect rela}
It is a new attempt to apply DRL methods to route planning. Through the interaction with the road network, the agent can learn the trends in traffic flow, get familiar with road network environment, and then train efficient navigation strategies according to specific requirements.

In this paper, we use DRL methods to optimize the agent's next action selection at each intersection, so as to minimize the total travel time.
At present, some studies have successfully applied DRL to solving traffic and transportation problems. Zolfpour-Arokhlo  \textit{et al.} \cite{rps1} proposed a system based on multi-agent reinforcement learning, in which each agent dealt with one of weather, traffic, road safety and fuel volume respectively. By learning the weights of the various components of the road network environment, the system developed prior route planning for vehicles and met the diverse needs of users. Yao, Peng and Xiao \cite{8456612} proposed a multi-objective optimization problem, considering both distance and safety. They designed a multi-objective heuristic algorithm based on reinforcement learning, and took the Pareto optimal solutions as alternative routes. Zhang \textit{et al.} \cite{9028238} proposed a novel control scheme for plug-in hybrid electric vehicles route planning. They designed a nonlinear approximator structure to approximate control actions and energy consumption, which successfully realized minimum energy cost after training.

On the other hand, different network structures can lead to different learning performance. Therefore, we need a reasonable network structure to enable agents to learn from the environment more effectively.
In DQN, the state-action value function was directly output by the network, where learning failed to generalize across actions. Therefore, Wang \textit{et al.} \cite{Duel} improved the structure of DQN. They split the value function into two parts, which were used for state evaluation and action advantage estimation respectively, and solved this issue.
Schaul \textit{et al}. \cite{ExperienceBuffer} designed a structure called Sum Tree to store serial numbers of samples in experience buffer. They no longer used the random method for sampling, but took samples according to the value of each sample, thus improving the learning efficiency.

Inspired by above works, we employ these DRL methods to achieve better learning performance, and propose a pedestrian route planning algorithm to minimize the total travel time while considering the change of pedestrian flow.

\section{System Model}\label{sect model}
\subsection{Road Network Model}\label{subsect_road_network}
We define the road network as a graph $\mathcal{G}(\mathcal{V},\mathcal{E})$, which includes the following parameters.
\begin{itemize}
	\item
	$\mathcal{V}=\{v_1,v_2,\dots,v_n\}$ represents the set of nodes, corresponding to intersection points in the road network.
	
	\item
	$\mathcal{E}=\{e_1,e_2,\dots,e_m\}$ represents the set of edges, corresponding to road segments connecting two nodes in $\mathcal{V}$.
	
	\item
	$\mathcal{D}$ is the distance matrix, represented by
	\begin{equation}\label{eq_matrix_d}
	\mathcal{D}=
	\begin{bmatrix}
	d_{11} & d_{12} & \dots & d_{1n} \\
	d_{21} & d_{22} & \dots & d_{2n} \\
	\vdots & \vdots & \ddots & \vdots \\
	d_{n1} & d_{n2} & \dots & d_{nn}
	\end{bmatrix}
	,
	\end{equation}
	where $d_{ij}$ denotes the distance between $v_i$ and $v_j$.
	
	\item
	$\mathcal{P}$ is the pedestrian flow matrix, represented by
	\begin{equation}\label{eq_matrix_p}
	\mathcal{P}=
	\begin{bmatrix}
	p_{11} & p_{12} & \dots & p_{1n} \\
	p_{21} & p_{22} & \dots & p_{2n} \\
	\vdots & \vdots & \ddots & \vdots \\
	p_{n1} & p_{n2} & \dots & p_{nn}
	\end{bmatrix}
	,
	\end{equation}
	where $p_{ij}$ represents the pedestrian volume from $v_i$ to $v_j$.
	
\end{itemize}
Note that, if there is no path between node $v_i$ and node $v_j$ (including if $i$ equals $j$), then $d_{ij}, p_{ij}, p_{ji}$ are all set to be -1.

To explore the relationship between pedestrian speed and density in grid maps, we adopt the well-known velocity-density exponential model proposed by Aoki \cite{Aoki}, to reflect the walking speed of individuals of different age groups and different types of people.
Accordingly, the velocity of pedestrians is expressed as
\begin{equation}\label{eq3_speed} x=x_0\rho_{ij}^{-0.8}, \end{equation}
where $x_0$ represents free pedestrian flow velocity which is considered as 1.34 m/s, and $\rho_{ij}={p_{ij}} / {d_{ij}}$ represents pedestrian density under the condition that road width is unit length.

We use $c_{ij}$ to represent the time cost from node $v_i$ to node $v_j$. With equation (\ref{eq3_speed}), $c_{ij}$ can be written as
\begin{equation}\label{eq3_1} c_{ij}=\frac{d_{ij}}{x_0\rho^{-0.8}_{ij}}=d_{ij}^{0.2}p_{ij}^{0.8}x_0^{-1}. \end{equation}
Road length $d_{ij}$ and pedestrian velocity $x_0$ are fixed, therefore $c_{ij}$ will increase with the increasement of pedestrian density $\rho_{ij}$, and the average passage time will then become longer.

\subsection{Pedestrian Flow Model}\label{subsect_pedestrian_flow}
We use time series methods to predict the change of pedestrian flow.
Then we can find the linear relationship between the number of people at the present moment and the number of people at the past several moments \cite{8556090,7905431}.
Considering the short-term correlation of pedestrian flow, we employ autoregressive integrated moving average (ARIMA) model, which is suitable for evaluating the dynamics and continuity of time series, and describing the relationship between past and present.

Denote $\nabla$ to be difference operator, we then obtain the expressions of the first-order and $d$-order difference, given by
\begin{equation}\label{eqarima_1} \nabla p_{ij}(t)=p_{ij}(t)-p_{ij}(t-1) , \end{equation}
\begin{equation}\label{eqarima_2} \nabla^d p_{ij}(t)=\nabla^{d-1} p_{ij}(t) - \nabla^{d-1} p_{ij}(t-1). \end{equation}

Denote $B$ to be delay operator which satisfies $y_{ij}(t-k)=B^ky_{ij}(t)$. Accordingly, we have $ \nabla^d=(1-B)^d$. Assuming the sequence $\{p_{ij}\}$ with $d$ time difference is stationary, we can express the corresponding ARIMA model as
\begin{equation}\label{eqarima_3} \lambda(B)\nabla^d p_{ij}(t) = \mu(B)\epsilon(t), \end{equation}
where $\epsilon(t)$ is white noise sequence that satisfies $\mathbb{E}[\epsilon(t)]=0, Var[\epsilon(t)]=\delta^2, \mathbb{E}[\epsilon(t)\epsilon(s)],s \neq t$.

Autoregressive coefficient polynomial and smoothing coefficient polynomial in equation (\ref{eqarima_3}) are
\begin{equation}\label{eqarima_4} \lambda(B)=1-\sum\limits_{k=1}^{O_1}\lambda_k B^k \end{equation}
and
\begin{equation}\label{eqarima_5} \mu(B)=1-\sum\limits_{k=1}^{O_2}\mu_k B^k \end{equation}
respectively, where positive integers $O_1$ and $O_2$ are order numbers for autoregressive model and moving average model.

Next we will combine existing data sets and generate time series of pedestrian flow through ARIMA model, and use the sequence for simulation setup.

\begin{figure*}[t]
	\centering
	\includegraphics[scale=0.8]{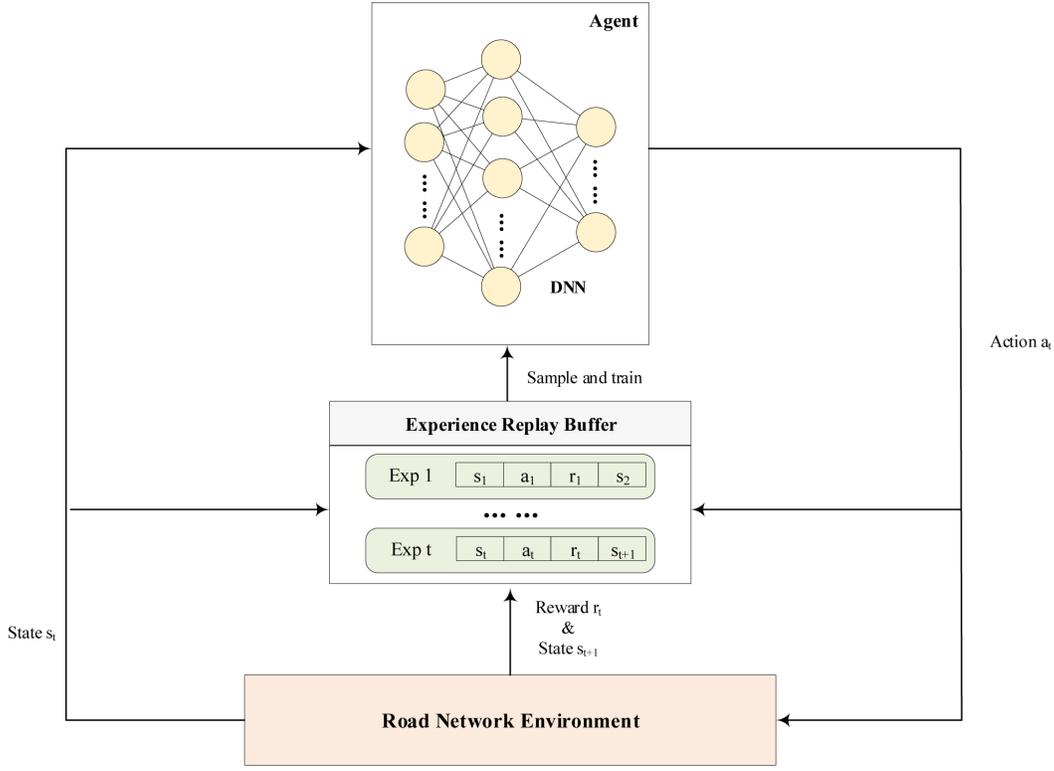}
	\caption{Diagram of DRL-based learning framework.}
	\label{DQN}
\end{figure*}

\section{Problem Formulation}\label{sect problem}
According to characteristics of the road network model, the environment exploration based on a path-finding strategy can be modeled as a discrete Markov decision process (MDP) to maximize the system reward.
In this section, we first describe the variables of MDP in route planning scenario, and then formulate our problem.

\subsection{Representation of State and Action}\label{subsect_problemA}
We model a regular road network as a grid map. Consider that each road intersection corresponds to a unique coordinate, we choose the agent's observation of current position as state $s_t$. For a grid map of size $N$ by $N$, its state space can be expressed as
\begin{equation}\label{eq_state} \mathcal{S}=\{loc_x\} \times \{loc_y\}, \end{equation}
where $loc_x, loc_y \in [0, N]$ .

In each time slot $t$, the agent selects an action $a_t$ according to the policy $\pi(a_t|s_t)$ before receiving a reward $r_t$ from the environment.
On the basis of the observation of current position, the agent can obtain a set of reachable nodes. Full action space for the agent can be expressed as
\begin{equation}\label{eq_action} \mathcal{A}=\{up, down, right, left\}\triangleq\{0,1,2,3\},  \end{equation}
where each element represents a different direction.

However, when the agent is at the edge of the map, its valid action space is actually a subset of $\mathcal{A}$. Note that, if the agent selects a direction and leave the area specified in current map, we determine that the selected action is invalid and will select another one from remaining directions.

\subsection{Reward and Optimization Problem}\label{subsect_problemB}
The goal of the agent is to find an optimal behavior policy to maximize the expected long-term discounted reward from each state $s_t$, \emph{i.e.}, to maximize the expectation of
\begin{equation}\label{eqRt} R_t=\sum\limits_{i=t}^T\gamma^{i-t}r_t, \end{equation}
where $T$ is the termination step, and $\gamma\in[0,1]$ denotes the discount factor that trades-off the importance of immediate and future rewards.

Every time the agent performs an action, it gets a feedback from the environment. We assume that the agent realizes the movement from node $v_i$ to node $v_j$ after the execution of a selected  action $a_t$, and the instantaneous reward function is then expressed as
\begin{equation}\label{eqreward} \begin{aligned}
r_t = -\frac{d_{ij}}{x_0\rho_{ij}(t)} + w_r \big(\phi(t)-\phi&(t-1)\big) \\
& + \textbf{1}(s_{t}=s_{end}).
\end{aligned}\end{equation}
The first term is the time that the agent takes to pass through the current road segment, which mainly determines the incentive for current action.
The second term is the motivation, which is used to motivate the agent to move toward the destination and to avoid meaningless wanderings during training. $\phi(t)$ represents the straight-line distance from current position to the destination in time slot $t$, and $w_r$ is a weight factor. In early stages of training, it is difficult for the agent to make a correct judgment when it is far from its destination. With this term, the agent can avoid choosing routes in the opposite direction under similar road conditions.
Note that, $w_r$ should be small in our problem, because time consumption is our main factor to consider.
The indicator function in the third term means that when the agent reaches the destination, it will be rewarded with an additional positive value.

In route planning, there could be several feasible paths in a graph. In this paper, we try to find the optimal one that takes the least amount of time to travel from origin node to destination node. 
Denote the complete action sequence of the agent by $\{a_1,a_2,\dots,a_T\}$. We then have the following optimization problem.
\begin{equation}\label{eqproblem} \begin{aligned}
\min\limits_{a_t} \ \mathbb{E}\Big[\sum\limits_{t=1}^{T}-\frac{d_{ij}}{x_0\rho_{ij}(t)} &+ w_r \big(\phi(t)-\phi(t-1)\big)\Big]  \\
s.t.\ \ a_t \in \mathcal{A}, \ &v_i\in \mathcal{V}, \ v_j \in \mathcal{V}
\end{aligned}\end{equation}

\section{Proposed Approach}\label{sect DRL}
The travel time is difficult to estimate, as pedestrian flow matrix $\mathcal{P}$ is changing all the time. In this section, we introduce a DRL based approach to tackle this issue.

Reinforcement learning develops a control policy by having agents interact with the environment, accumulating and learning from experience of failures or successes. In order to get the accurate value of $R_t$ in (\ref{eqRt}), we employ the widely-used action-value function to represent the expected return after selecting action $a_t$ in state $s_t$ according to policy $\pi$.
\begin{equation}\label{eq4_2} Q^{\pi}(s_t,a_t)=\mathbb{E}[R_t|s_t,a_t,\pi], \end{equation}
which can be calculated by using the recursive relationship given by the Bellman equation:
\begin{equation}\label{eq4_3} Q^{\pi}(s_t,a_t)=\mathbb{E}\big[r_t + \gamma\mathbb{E}[Q^{\pi}(s_{t+1},a_{t+1})]\big],  \end{equation}
and Q-values are updated recursively by using temporal difference method:
\begin{equation}\label{eq4_4}\begin{aligned}
Q^{\pi}(s_t,a_t)\leftarrow &Q^{\pi}(s_t,a_t)+\alpha[r_t+ \\
&\gamma\max\limits_{a_{t+1}}Q^{\pi}(s_{t+1},a_{t+1})-Q^{\pi}(s_t,a_t)].
\end{aligned}\end{equation}

To deal with the complex problem with large state space, we employ deep Q network that combines RL methods with DNN. Neural network is employed as a nonlinear function approximator to estimate the action-value function $Q(s,a;\theta) \approx Q^{\pi}(s,a)$. Our DRL-based learning framework is shown in Fig. \ref{DQN}. First, the agent observes the state and outputs an action. Then the environment feeds back a reward and enters the next state. These elements constitute an experience tuple $e_t=(s_t,a_t,r_t,s_{t+1})$ at time step $t$. Finally, the agent samples a batch of memories from experience replay buffer and calculates a set of loss functions below.
\begin{equation}\label{eq4_5} L_i(\theta_i)=\mathbb{E}_{e_t\sim\mathcal{B}}[(y_i-Q(s_t, a_t; \theta_i))^2], \end{equation}
with
\begin{equation}\label{eq4_6} y_i=r_t + \gamma \max\limits_{a_{t+1}}Q^{\pi}(s_{t+1}, a_{t+1}; \theta^{-}), \end{equation}
where $\theta^{-}$ is the old network parameter, and is replaced by $\theta_{i-1}$ from the evaluate network after a period of time.

In order to reflect the difference in value of the different actions, we further use dueling network to separate the above Q function into two parts, \textit{i.e.}, the valuation of current state and the advantage of choosing different actions.
\begin{equation}\label{eq_duel_1} Q(s_t,a_t;\theta,w_1,w_2)=V(s_t;\theta,w_1)+A(s_t,a_t;\theta,w_2), \end{equation}
where $Q(s_t,a_t;\theta,w_1,w_2)$ is actually another form of $Q(s_t,a_t;\theta)$, and $\theta$ is the parameter of the common part in neural network architecture. $w_1$ and $w_2$ are newly introduced parameters, which denote parameters of fully-connected layers for state valuation and action advantage evaluation respectively.

Note that the expectation of $Q(s_t,a_t)$ is $V(s_t)$, we have $\mathbb{E}_{a\sim\pi}[A(s_t,a_t)]=0$. Therefore, we turn the second part of (\ref{eq_duel_1}) into the advantage of the current action's evaluation compared to the average reward of all actions.
\begin{equation}\label{eq_duel_2} \begin{aligned}
Q(s_t,a_t;&\theta,w_1,w_2)=V(s_t;\theta,w_1) + \\
&\Big(A(s_t,a_t;\theta,w_2)-\frac{1}{|A|}\sum\limits_{a^{\prime}}A(s_t,a^{\prime};\theta,w_2)\Big),
\end{aligned}\end{equation}
where $|A|$ denotes the number of valid actions.

In training process, we perform difference operation on loss functions in equation (\ref{eq4_5}), and obtain a batch of gradient.
\begin{equation}\label{eq4_7} \begin{aligned}
\nabla_{\theta_i}L_i(\theta_i)=\mathbb{E}_{e_t\sim\mathcal{B}}\Big[(y_i-Q&(s_t,a_t;\theta_i,w_1,w_2)) \\ &\nabla_{\theta_i}Q(s_t,a_t;\theta_i,w_1,w_2)\Big].
\end{aligned}\end{equation}

Then we use the following standard non-centered RMSProp optimization algorithm \cite{RMSProp} to update parameters in the neural network.
\begin{equation}\label{eq4_8} \theta\leftarrow\theta-\eta\frac{\Delta\theta}{g+\sigma}, \end{equation}
with
\begin{equation}\label{eq4_9} g=\alpha g +(1-\alpha)\Delta\theta^2. \end{equation}

We present the process of our dueling deep Q network based learning method. Please refer to Algorithm \ref{algoDDQN} for detailed steps of the DARP algorithm.
\begin{algorithm}[htb]
	\caption{Dueling Deep Q Network Based Dynamically Adjustable Route Planning (DARP)}
	\label{algoDDQN}
	\begin{algorithmic}[1]
		\STATE Initialize prioritized experience replay buffer $\mathcal{B}$.
		\STATE Initialize Q network with random weights $\theta$.
		\STATE Initialize target Q network with $\theta^{-}=\theta$.
		\FOR{episode $epi=1,2,\dots,epi_{max}$}
		\STATE Initialize the environment, get initial position $s_1$.
		\FOR{time slot $t=1,2,\dots,t_{max}$}
		\STATE Choose action $a_t$ using epsilon-greedy method with a fix parameter $\epsilon$.
		\STATE Go in the direction indicated by $a_t$, obtain reward $r_t$ from environment and observe next position $s_{t+1}$.
		\STATE Collect experience and save the tuple $e_t$ in $\mathcal{B}$.
		\STATE Sample a batch of experiences $(s_j,a_j,r_{j},s_{j+1})$ from $\mathcal{B}$.
		\IF {episode terminates at time slot $j+1$}
		\STATE Calculate $y_j$ according (\ref{eq4_6});
		\ELSE
		\STATE Set $y_j=r_t$.
		\ENDIF
		\STATE Perform gradient descent and update Q-network according to (\ref{eq4_7}).
		\STATE Update current state, and every $K$ steps set $\theta^{-}=\theta$.
		\ENDFOR
		\ENDFOR
	\end{algorithmic}
\end{algorithm}

\section{Evaluation}\label{sect result}
In this section, we first describe the simulation setup, and then carry out experiments to evaluate our algorithm.

\subsection{Experiment Setup}\label{subsect_setup}
We build experiments on a $5 \times 5$ virtual grid map, which contains 36 nodes and 60 edges. At the same time, we generate the distance matrix $\mathcal{D}$ given by (\ref{eq_matrix_d}), where the distance between any two adjacent nodes is randomly generated from a range of 100 m to 1 km. 

Note that the distance matrix is fixed in our environment, while the pedestrian flow matrix is changing all the time. To model the dynamic environment, we take the following steps.
\begin{itemize}
	\item \textbf{Step 1: Analysis.}
	Based on existing data sets, we fit the change curve of pedestrian flow in Shanghai Zoo by employing ARIMA model with $O_1=1$ and $O_2=0$. Then we obtain the relationship between pedestrian flow and time in real environment.
	
	\item \textbf{Step 2: Initialization.}
	Initial value of pedestrian flow on each edge is randomly generated from a range of 200 to 1000. In addition, we randomly select several edges as crowded road segments in part of experiments, and their initial values are randomly generated and follow a Gaussian distribution with mean of 5000 and variance of 1000.
	
	\item \textbf{Step 3: Prediction.}
	The trained ARIMA model is used to predict the number of pedestrians in each road segment in the future. In the interaction between the agent and the environment, these time series will be used as the basis for updating the pedestrian flow matrix $\mathcal{P}$ given by (\ref{eq_matrix_p}).
\end{itemize}

\begin{figure}[ht]
	\centering
	\includegraphics[scale=0.935]{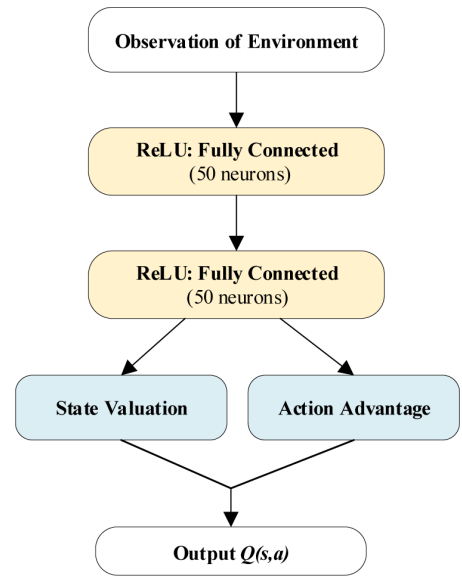}
	\caption{DNN architecture.}
	\label{dnn}
\end{figure}
To implement the DRL architecture, two separate deep neural networks are used, both of which share the same structure. Several fully-connected layers are employed in our neural network, as shown in Fig. \ref{dnn}. The number of neurons in input and output layer corresponds to the dimensions of state and action, respectively. Each of the two hidden layers has 50 neurons, and ReLU function is employed as activation function. We set replay buffer size to be $10^4$ and minibatch memory size to be 32. Discount factor used in Q-learning update is 0.95 and learning rate used by RMSProp is $10^{-4}$. In addition, the weight factor $w_r$ in reward function is set to be $10^{-3}$.

We use the following baseline methods to compare with our DARP algorithm.

\textbf{Random Selection}: At each moment, the agent randomly selects a node as the moving target.

\textbf{Shortest Path Selection}: The agent moves along the shortest path according to shortest path planning methods, \textit{e.g.},  A* algorithm.

\subsection{Numerical Results}\label{subsect_simulation}
Our experiment is completed in Pycharm IDE, and the TensorFlow version used is 1.15.0. All the experimental results are drawn using Python Matplotlib Library.

\begin{figure}[ht]
	\centering
	\includegraphics[scale=0.37]{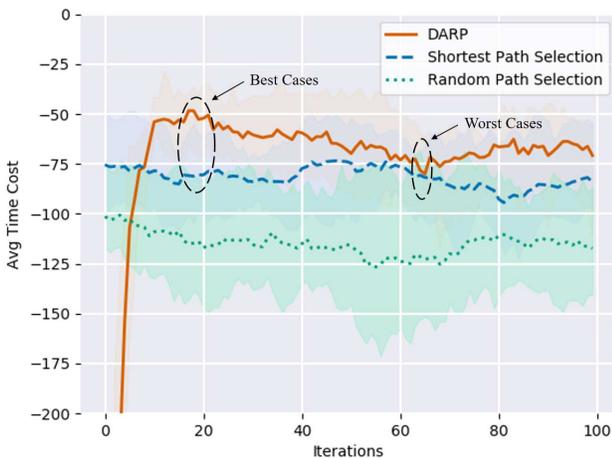}
	\caption{Time cost of different methods under unblocked conditions.}
	\label{unblocked}
\end{figure}

We first set the initial state to be non-congested, where the pedestrian flow at each edge is uniformly distributed. We evaluate the performance of different methods, and depict the result in Fig. \ref{unblocked}, where each method is tested 10 times and the average performance is recorded.
The horizontal axis represents the number of training rounds, and the vertical axis represents the cumulative time cost of the agent from original node to destination node.
Note that in DRL architecture, we take the negative form of time consumption as a reward. Therefore, the closer the 'Avg Time Cost' is to 0, the better the performance.

DRL based approach usually has a lower reward at the beginning of training, because during this stage, the agent is trying out all possible paths on the map and collecting initial experience. After about 10 iterations of training, the average time cost obtained by our approach reaches convergence at about -67 and then fluctuates slightly in a small range. We can easily find that the random path selection scheme has the worst performance. Compared with the shortest path selection scheme, our approach can save up to $37.3\%$ of the time cost in the best cases.
In the worst cases, our approach can still exhibit better performance.
This is because under conditions that all roads are generally free of congestion, the shortest path is usually the one that consumes the least time, and the detour will lead to longer time cost due to the increase of total distance.
However, the unblocked state at the beginning does not mean that it will remain so thereafter. In this dynamic process, congestion may still occur in the future, and our approach will still be able to preserve the best performance among all methods.

\begin{figure}[ht]
	\centering
	\includegraphics[scale=0.37]{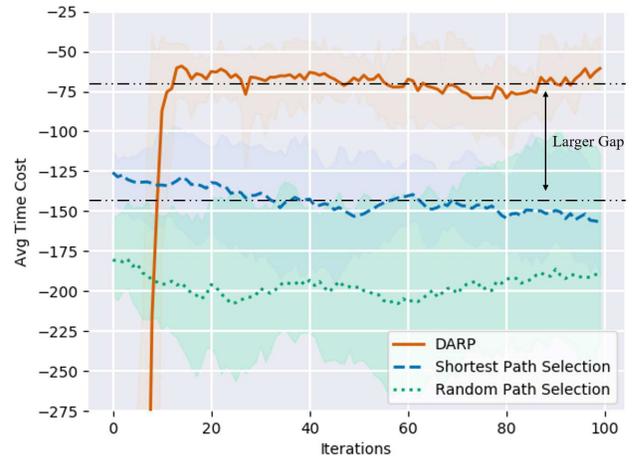}
	\caption{Time cost of different methods under congested conditions. }
	\label{congested}
\end{figure}

We further explore the performance of our algorithm under local congestion. We randomly select several nodes, and the edges connected to them are set as crowded road segments which have high values of pedestrian flow. Fig. \ref{congested} shows that the DARP algorithm converges after about 15 iterations and then save more travel time for users.
Compared with the shortest path selection scheme and random path selection scheme, our approach can save $52.1\%$ and $65.3\%$ time, respectively.

From Figs. \ref{unblocked} and \ref{congested}, we also observe that although some road segments become congested, the time cost obtained by our DARP algorithm does not change much, maintaining about -75, while the time cost of the other baseline methods are greatly increased. This is because by constantly interacting with the environment, the agent gradually learns about the crowded areas in the map and grasps the changing rules of pedestrian flow. Therefore, the agent can always choose a better path after trial and error, bypassing congested areas.
However, the other two baseline methods do not have the ability to dynamically adjust their stategies, and thus the rewards are considerably reduced, \textit{i.e.}, the time taken to complete the trip increases dramatically.
Once congestion road segments occur on routes provided by these methods, then the total travel time will greatly increase.

\section{Conclusion and Future Works}\label{sect Conclusion}
In this paper, we propose a route planning algorithm based on deep reinforcement learning to minimize the total travel time. 
Unlike previous studies, our approach focuses on pedestrians and does not require any prior knowledge of road network.
The agent interacts with the environment, acquires needed information and learns corresponding strategies by itself.
Therefore, the proposed method does not require human involvement, so it can adapt to different environments, even if the road network environment changes. 
Simulation results show that our method can save $52\%$ time cost in the case of congestion, and also has better performance than the traditional methods under unblocked conditions.

Note that, our model only considers the pedestrian, and ignores the impact of vehicles and traffic lights on pedestrians' movement.
Under these circumstances, pedestrians may sometimes be forced to stop moving, and the corresponding traffic flow model will be more complex.
In the future work, we would like to consider these factors, and explore the route planning problem under larger maps.

\bibliographystyle{IEEEtran}
\bibliography{paper_ref}

\end{document}